%% file: main.tex
\documentclass{article} %
\usepackage{iclr2027_conference,times}

\input{math_commands.tex}

\usepackage[pagebackref=true,breaklinks=true,colorlinks=true,urlcolor=magenta,bookmarks=false,citecolor=blue]{hyperref}
\usepackage{url}
\usepackage{booktabs}
\usepackage{multirow}
\usepackage{arydshln}
\usepackage{placeins}
\usepackage{graphicx}
\usepackage{capt-of}
\usepackage{microtype}
\graphicspath{{figs/}}
\usepackage{amsmath,amssymb,amsthm}

\newcommand{\factorsplat}{FactorSplat\xspace}
\newcommand{\tf}{T}
\newcommand{\tfbase}{T_{0}}
\newcommand{\zitf}{\mathbf{z}_{i,\tf}}

\newcommand{\deltamet}{\Delta_{\mathrm{TF}}}
\usepackage{xspace}

\title{\factorsplat{}: Appearance-Controllable Gaussian Proxies for Medical Volume Rendering}

\author{Zhongpai Gao\thanks{Corresponding author}, Benjamin Planche, Meng Zheng, Anwesa Choudhuri, \\
\textbf{Terrence Chen, Ziyan Wu} \\
United Imaging Intelligence, Boston MA, USA
}

\iclrfinalcopy
\begin{document}

\maketitle
\lhead{Preprint}

\vspace{-0.7em}
\begin{minipage}{\textwidth}
\centering
\includegraphics[width=\textwidth]{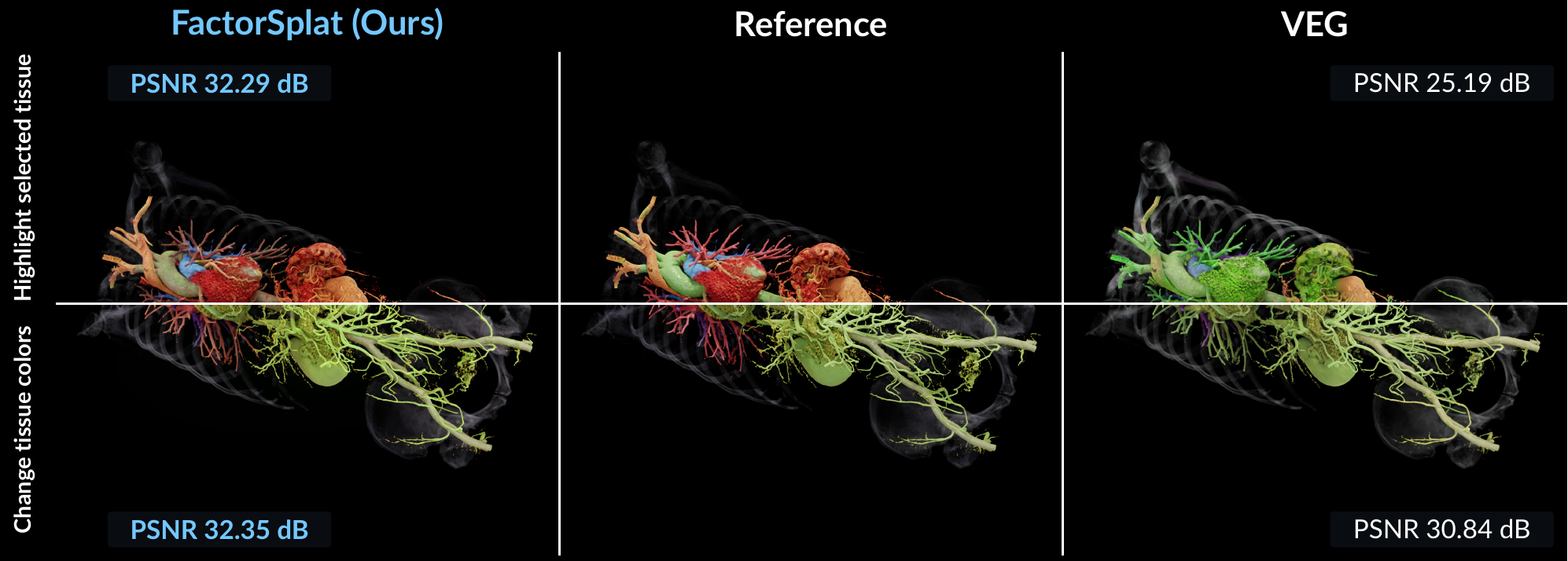}
\captionof{figure}{\textbf{Change appearance without retraining.} Two vascular TF edits:
highlight selected tissue (top) and change tissue colors (bottom).
Full-frame PSNR.}
\label{fig:teaser}
\end{minipage}
\vspace{0.4em}

\begin{abstract}
Transfer functions (TFs) control color and visibility in medical volume rendering,
but image-trained Gaussian proxies typically bake one transfer function into
their appearance. We present \factorsplat{}, a per-scene N-dimensional Gaussian
splatting (N-DGS) proxy that accepts region-specific intensity-to-RGBA curves
at inference. A local lookup applies the authored color and opacity change,
while a shared functional encoder and low-rank per-Gaussian factors learn
the residual appearance response. Geometry and directional appearance remain
shared across presets, with visibility control and TF-aware pruning preserving
the ability to hide and reveal structures. On seven CT and MR scans, FactorSplat improves mean PSNR and
changed-region error over region-aware VEG across validation, interpolation,
unseen composition, and out-of-distribution (OOD) edits. Across these four
splits, seven-scan mean PSNR gains over VEG range from 1.10 to 1.52~dB.
One checkpoint per scan supports unseen edits without retraining.
At $1600^2$, the cached fast renderer averages 524 FPS with 1.17~ms TF switches.
The project page is: \url{https://gaozhongpai.github.io/FactorSplat/}.
\end{abstract}

\section{Introduction}
\label{sec:intro}
Transfer functions (TFs) control the appearance of volume renderings by mapping
measured intensity and derived attributes to color and opacity. They can
suppress one material, expose another, or separate structures that overlap
in a one-dimensional histogram~\citep{kniss2002multidimensional}.  In the
clinical renderer considered here, each segmented region has an authored
intensity-to-RGBA curve (HU for CT).  Segmentation selects the appropriate
curve, whose resulting color and opacity govern volumetric transport.  A
preset change can therefore alter both which anatomy contributes to the image
and how the visible anatomy occludes, shades, and transmits light in cinematic
rendering~\citep{dappa2016cinematic}.

Image-trained Gaussian proxies offer an efficient representation of such
renderings. A proxy optimized from high-quality
volume renderings is compact and rasterizes in real time
~\citep{niedermayr2024application,gao2025renderfm}. These fixed-appearance proxies encode
the training appearance in per-Gaussian spherical-harmonic coefficients and
opacities, so the TF is no longer an input. Supporting another
preset then requires rebuilding or replacing the proxy. Moreover, training on one opaque
preset may omit anatomy that a later preset should reveal.

Making the TF an inference-time control poses three coupled problems. First, the
condition is a \emph{function}, not a categorical preset identifier: a model
should respond to a new sampled curve without allocating new parameters.
Second, the response is spatially heterogeneous. The same TF edit should
affect a Gaussian according to the region curves and intensities it covers. Third,
training and densification must preserve \emph{coverage} across the condition
family so that structures revealed by any supported preset remain represented.
Because a TF changes optical appearance rather than anatomy, geometry should
remain invariant to this control.

We address these challenges with \factorsplat{}, a per-scene Gaussian proxy
trained across TFs. Gaussian centers, covariances, and the N-DGS view response
are shared. Local region--intensity samples drive two complementary appearance
branches: a physical lookup applies the TF difference directly, while a shared
pointwise encoder and low-rank per-Gaussian factors model the cinematic
renderer's residual response. The lookup anchors the prediction to the authored
edit; the learned branch corrects its image-space effect.
Both branches modify only the view-independent (DC) color and opacity; directional SH
coefficients remain TF-independent.  Centering the branches on an authored
base TF gives exact identity at that preset, and pruning by maximum opacity
over the training bank protects primitives needed by different training TFs.

Our main contributions are:
\begin{itemize}
  \item \textbf{Representation:} FactorSplat conditions an N-DGS proxy on
  region-specific intensity-to-RGBA curves, rendering unseen TFs with shared
  geometry and no refitting.
  \item \textbf{Appearance model:} A hybrid model combines an interpretable
  local TF lookup, a low-rank DC-color/opacity residual, analytic visibility
  control, and TF-aware coverage preservation.
  \item \textbf{Evaluation:} Across seven scans, one checkpoint per scan
  improves mean PSNR and edit fidelity over region-aware VEG on unseen TFs.
\end{itemize}

\section{Related Work}
\label{sec:related}
\noindent\textbf{Transfer functions and learned visualization synthesis.}
Multidimensional TFs use intensity, gradients, or other attributes to classify
overlapping volume structures and assign optical properties
interactively~\citep{kniss2002multidimensional}. A complementary line of work
helps users \emph{choose} that mapping: differentiable direct volume rendering
optimizes TF parameters against an image objective~\citep{weiss2022diffdvr},
and recent semantic systems derive mappings from selected structures or text
prompts~\citep{engel2024segmentationtf,jeong2024texttf}. FactorSplat reproduces
the rendering of an authored TF with a
compact proxy.

Image-space visualization surrogates condition neural generators on viewpoint
and rendering parameters. \citet{berger2019generative} encode
color and opacity TFs into a view-invariant latent space, while
DNN-VolVis~\citep{hong2019dnnvolvis} predicts volume-rendered images for
interactive exploration. These models can learn the output of either local or
global-illumination renderers, but an image generator has no explicit 3D
primitive representation. ViSNeRF introduces a factorized radiance field for
interpolation over TFs, isovalues, time, and simulation parameters
~\citep{yao2025visnerf}; it retains ray-based field rendering rather than an
explicit rasterizable proxy. Our functional encoder likewise keeps rendering
controls explicit, but its output modulates a per-scene Gaussian field.

\noindent\textbf{Gaussian proxies for volume rendering.}
3D Gaussian proxies trained on multi-view volume renderings enable compact,
real-time viewing of cinematic anatomy~\citep{niedermayr2024application};
Render-FM amortizes proxy construction across scans by predicting an N-DGS
representation feed-forward~\citep{gao2025renderfm}. Both target a fixed
rendering appearance. iVR-GS instead trains editable basic models for disjoint
intensity ranges and composes them to control color, opacity, and Blinn--Phong
lighting~\citep{tang2025ivrgs}. Its cost and representational size grow with
the number of basic models. ECoNGS shares parameters and compresses anchors across basic
scenes, reducing redundancy while retaining the composition
model~\citep{tang2026econgs}.

VEG is the closest transfer-function-agnostic Gaussian representation
~\citep{dyken2026veg}. It initializes Gaussians from structured or unstructured
volume data, stores scalar values rather than baked color and opacity, and
trains under a set of opacity maps that covers the scalar range. At inference,
it directly applies unseen one-dimensional color and opacity mappings under an
emission--absorption renderer with Blinn--Phong shading. This factorization
directly links Gaussian appearance to the scalar field. Our setting instead
uses region-specific intensity TFs and cinematic-rendering supervision. We
combine direct lookup with an image-trained N-DGS proxy whose residual learns
the TF-dependent appearance response. Section~\ref{sec:experiments} describes
our region-aware VEG extension.

\noindent\textbf{Conditioned explicit Gaussian fields.}
N-dimensional Gaussian splatting has conditioned primitives on viewing
direction and time~\citep{gao20246dgs,gao20257dgs,liu2025universal}.
\factorsplat{} uses the opacity-only N-DGS formulation as its view-dependent
backbone and adds a separate TF axis only to appearance. Concurrent scientific
visualization systems also condition Gaussian attributes. GS-Surrogate learns
sequential simulation- and visualization-parameter deformations and disables
geometry heads for its opacity-only TF experiment~\citep{li2026gssurrogate};
ParticleGS maps global particle radius and opacity controls to Gaussian scale
and opacity~\citep{jiang2026particlegs}. These methods condition Gaussian
fields on low-dimensional controls. DeepONet~\citep{lu2021deeponet}
encodes input functions for a separable output representation. FactorSplat
instead grounds function-valued TF inputs in each Gaussian's local support,
with per-primitive factors mapping local codes to appearance offsets rather
than generating full weights with a hypernetwork~\citep{ha2017hypernetworks}.

\section{Method}
\label{sec:method}

\begin{figure*}[t]
\centering
\includegraphics[width=\textwidth]{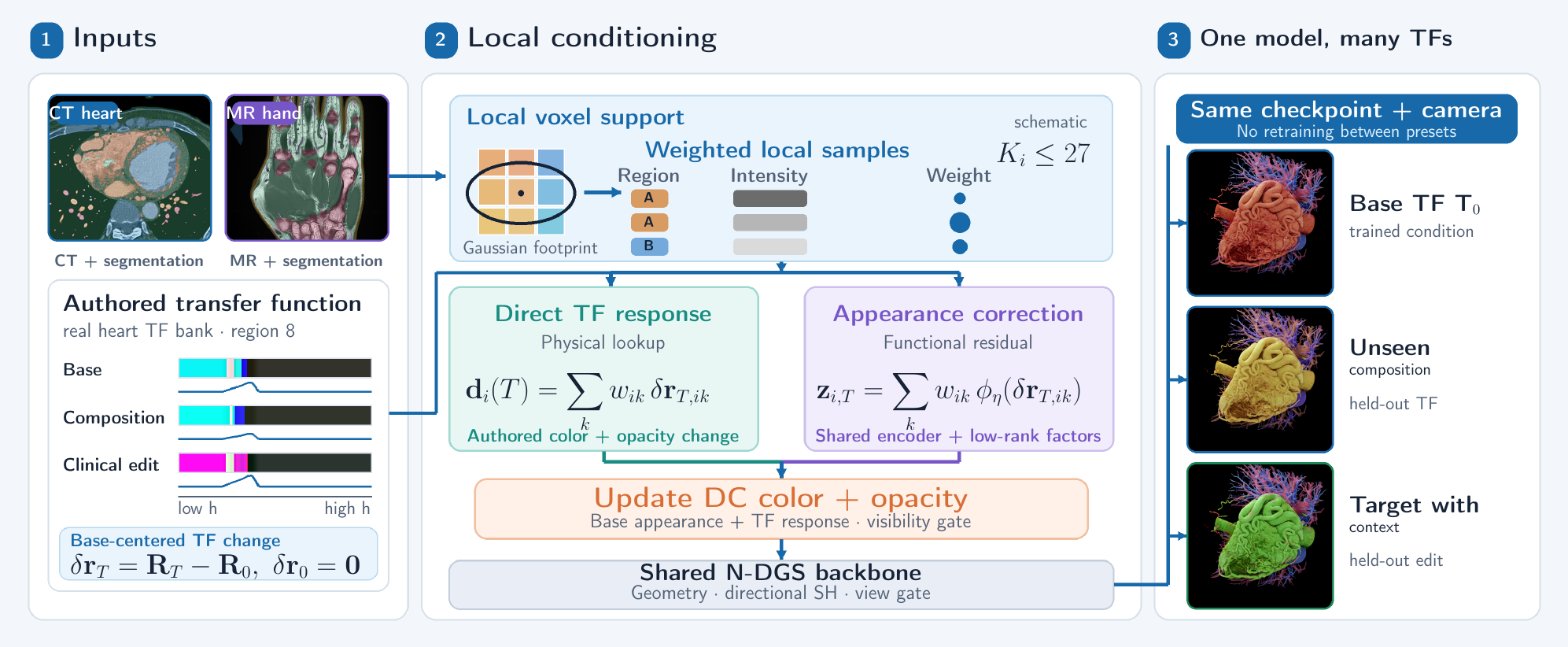}
\caption{\textbf{Method overview.} FactorSplat combines a physical lookup and
low-rank residual over local material--intensity support to condition DC color
and opacity.}
\label{fig:method-overview}
\vspace{-1em}
\end{figure*}

Figure~\ref{fig:method-overview} separates shared scene structure from
TF-dependent appearance. Local material descriptors connect authored curves
to Gaussian color and opacity through a direct lookup and a learned residual.
TF changes update these attributes once; camera navigation reuses them.

\subsection{Problem formulation and N-DGS backbone}
\label{subsec:setting}

A scene contains $N$ Gaussian primitives rasterized by EWA splatting and
front-to-back alpha compositing~\citep{zwicker2001ewa,3dgs}.  For primitive
$i$, $\bm\mu_i\in\mathbb R^3$, $\bm\Sigma_i\in\mathbb R^{3\times3}$, and
$o_i\in\mathbb R$ denote its center, covariance, and opacity logit; its base
opacity is $\sigma(o_i)$ for logistic sigmoid $\sigma$. Every N-DGS-based
model in our comparison uses degree-one spherical harmonics: $Q_b=4$ basis functions, one DC
and three directional. Let
$\mathbf C_i^{\mathrm{base}}\in\mathbb R^{Q_b\times3}$ collect their RGB
coefficients and $Y_q(\boldsymbol\omega_i(\pi))$ denote basis $q$ evaluated at
the viewing direction from primitive $i$ to camera $\pi$.  Before the usual
rasterizer activation, the shared backbone color is
$\sum_{q=0}^{Q_b-1}Y_q(\boldsymbol\omega_i(\pi))
\mathbf C_{i,q}^{\mathrm{base}}$, with $q=0$ denoting DC.

We use an opacity-only N-DGS backbone~\citep{gao20246dgs}, with directly
parameterized view-dependent opacity and no view-dependent position shifts.
Let $\pi$ denote a camera and
$s_i^{\mathrm{view}}(\pi)\in(0,1]$ the TF-independent opacity attenuation
produced by N-DGS for primitive $i$.  The backbone opacity passed to the
rasterizer is $\sigma(o_i)s_i^{\mathrm{view}}(\pi)$.  \factorsplat{} leaves this
gate unchanged and keeps all geometry shared across TFs:
$\bm\mu_i(\pi,\tf)=\bm\mu_i$ and
$\bm\Sigma_i(\pi,\tf)=\bm\Sigma_i$.

A preset contains one intensity-to-RGBA \emph{transfer curve} for every segmented
renderer region.  Writing $m$ for the region/material channel and $h$ for the
calibrated scalar intensity,
\begin{equation}
\mathbf R_{\tf}(m,h)
=\big(\boldsymbol\rho_{\tf}(m,h),a_{\tf}(m,h)\big)
\in[0,1]^3\times[0,1].
\label{eq:tf}
\end{equation}
Here $h$ is calibrated HU for CT and the renderer's normalized scalar for MR;
$m\in\{0,\ldots,M-1\}$ indexes one of $M$ region curves,
$\boldsymbol\rho_{\tf}$ is RGB, and $a_{\tf}$ is transfer opacity.  The index
$m$ routes a voxel to its authored curve; the encoder consumes the curve's
RGBA values rather than the region identifier.
Exporting every preset on the common grid $\{h_s\}_{s=1}^{S}$ gives the
sampled input $\mathbf R_{\tf}\in[0,1]^{M\times S\times4}$;
binary region-visibility flags are denoted separately by $v_{\tf}$ below.
We designate one authored preset $\tfbase$ as the base condition.

Given reference images for a training bank $\mathcal{T}_{\mathrm{train}}$, we
jointly learn one proxy that renders unseen $\tf\notin\mathcal{T}_{\mathrm{train}}$
without optimization or per-preset parameters.

\subsection{Local transfer-function conditioning}
\label{subsec:conditioning}

A Gaussian fitted to cinematic images combines material response, shading,
and visibility. We therefore separate the authored TF change from its effect
on the fitted proxy. The physical lookup supplies a local color and opacity
offset; the learned residual corrects the primitive's DC color and opacity
for the same edit. Geometry and directional appearance are shared across TFs.

\noindent\textbf{Local material--intensity descriptor and physical lookup.}
Let $\mathbf R_{\tf}^{(B)}\in[0,1]^{M\times B\times4}$ denote the raw TF curves
averaged onto $B=64$ intensity bins.  At initialization, we sample the $3^3$
voxel window around primitive $i$.  Its $k$th valid sample supplies a region
channel $m_{ik}\in\{0,\ldots,M-1\}$ and an intensity-bin index
$b_{ik}\in\{0,\ldots,B-1\}$; $K_i\leq K=27$ is the number of valid samples.
Normalized nonnegative weights $w_{ik}$ satisfy
$\sum_{k=1}^{K_i}w_{ik}=1$ and define the empirical joint mass
$p_i(m,b)=\sum_k w_{ik}\mathbf 1[(m_{ik},b_{ik})=(m,b)]$.
Both conditional branches read the same quantity, the raw,
unpremultiplied RGBA change that the preset applies to one
region--intensity entry,
\begin{equation}
\delta\mathbf r_{\tf}(m,b)=
\mathbf R_{\tf}^{(B)}(m,b)-
\mathbf R_{\tfbase}^{(B)}(m,b)\in\mathbb R^4 .
\label{eq:local-delta}
\end{equation}
Keeping the channels unpremultiplied separates color and opacity offsets. The physical
\emph{local lookup} is the weighted mean of this change over the primitive's own
samples,
\begin{equation}
\mathbf d_i(\tf)=\sum_{k=1}^{K_i}w_{ik}
\delta\mathbf r_{\tf}(m_{ik},b_{ik}).
\label{eq:lookup}
\end{equation}
Here $\mathbf d_i=(\mathbf d_i^c,d_i^\alpha)\in\mathbb R^3\times\mathbb R$
and $\mathbf d_i=\mathbf0$ when $K_i=0$.  We encode each pair as
$u_{ik}=m_{ik}B+b_{ik}$. The region index only selects the corresponding TF
curve: permuting the curve axis and identifiers together leaves the lookup
unchanged. Material associations persist during training
(Section~\ref{subsec:training}).

\noindent\textbf{Local functional encoder.}
The learned branch embeds each RGBA change before weighted, permutation-invariant
pooling, following the set-encoding principle of Deep Sets~\citep{zaheer2017deepsets}.
A bias-free pointwise MLP
$\phi_\eta(\mathbf v)=W_2\operatorname{ReLU}(W_1\mathbf v)$, with
$W_1\in\mathbb R^{H\times4}$ and $W_2\in\mathbb R^{r\times H}$, maps each
RGBA difference to a shared response basis. Primitive $i$ pools its local codes:
\begin{equation}
\zitf=\sum_{k=1}^{K_i}w_{ik}
\phi_\eta\!\left(
\delta\mathbf r_{\tf}(m_{ik},b_{ik})\right)
\in\mathbb R^r,
\label{eq:encoder}
\end{equation}
with $\zitf=\mathbf0$ when $K_i=0$. The MLP is evaluated once over the
$M\times B$ table at a TF switch; packed gathers then produce all primitive
codes. Each code depends on the primitive's own support, allowing spatially
different responses to the same preset. The shared RGBA encoder accepts
unseen TFs without preset-specific parameters. Centering and bias-free layers ensure
$\mathbf z_{i,\tfbase}=\mathbf0$ exactly.
Encoding before pooling retains responses to individual local edits that
could cancel if their RGBA differences were averaged first.
For example, changing a bone curve contributes only to codes whose descriptors
include bone, while soft-tissue edits act through their own support.

\subsection{Factorized appearance response}

\noindent\textbf{Low-rank per-Gaussian factors.}
Each primitive carries color factors $\mathbf A_i^{c}\in\mathbb{R}^{3\times r}$
and opacity factors $\mathbf a_i^{\alpha}\in\mathbb{R}^{r}$.  They translate
the local code into a DC-SH residual $\mathbf r_i^c\in\mathbb R^3$ and an
opacity-logit residual $r_i^\alpha\in\mathbb R$.  Define the authored
alpha-change indicator
\begin{equation}
q^\alpha(\tf)=
\mathbf{1}\!\left[
\left\|\mathbf R_{\tf}^{(B)}[:,: ,\alpha]
-\mathbf R_{\tfbase}^{(B)}[:,: ,\alpha]\right\|_\infty>0\right].
\label{eq:alpha-identity}
\end{equation}
The factor responses are
\begin{equation}
\mathbf r_i^c(\tf)=\lambda_R^c\tanh(\mathbf A_i^c\zitf),\qquad
r_i^\alpha(\tf)=q^\alpha(\tf)\lambda_R^\alpha
(\mathbf a_i^\alpha)^\top\zitf .
\label{eq:lowrank}
\end{equation}
Here $\lambda_R^c,\lambda_R^\alpha>0$ are fixed output scales and $\tanh$ acts
elementwise. For RGB-only edits, $q^\alpha=0$ keeps opacity unchanged; authored
alpha edits enable the learned opacity response. Stacking all
color and opacity factors yields a $4N\times r$
matrix. The latent dimension $r$ controls the size of the shared response basis
before the output nonlinearity, independently of the number of TFs.
The factors, shared encoder, and base appearance are jointly optimized.

\noindent\textbf{Complete appearance model.}
Let $b_L,b_R\in\{0,1\}$ enable the lookup and residual branches, respectively,
and let $\mathbf g^c\in\mathbb R^3$ and $g^\alpha\in\mathbb R$ be the learned
global lookup gains.  With fixed positive channel scales
$\lambda_L^c,\lambda_L^\alpha,\lambda_R^c,\lambda_R^\alpha$, we combine both
branches as
\begin{align}
\mathbf C_i(\tf)
 &=\mathbf C_i^{\mathrm{base}}
 +\mathbf e_0\!\left[
   b_L\frac{\lambda_L^c}{C_0}
   \big(\mathbf g^c\!\odot\!\mathbf d_i^c(\tf)\big)
   +b_R\mathbf r_i^c(\tf)\right]^{\!\top}, \label{eq:complete-color}\\
\bar\alpha_i(\tf)
 &=\sigma\!\left(o_i
 +b_L\lambda_L^\alpha g^\alpha d_i^\alpha(\tf)
 +b_R r_i^\alpha(\tf)\right), \label{eq:complete-opacity}\\
\alpha_i(\pi,\tf)
 &=\gamma_i(\tf)\,\bar\alpha_i(\tf)\,
 s_i^{\mathrm{view}}(\pi),
 \label{eq:complete-view}
\end{align}
where $\mathbf e_0=(1,0,0,0)^\top\in\mathbb R^{Q_b}$ selects the DC row,
$C_0=1/(2\sqrt{\pi})$ converts RGB offsets to DC-SH coefficients,
$\odot$ denotes componentwise multiplication, and $\bar\alpha_i(\tf)$ is the
TF-conditioned opacity before the shared N-DGS view gate. The settings
$(b_L,b_R)=(1,0),(0,1),(1,1)$ define the
\emph{lookup}, \emph{residual}, and \emph{hybrid} variants, respectively.
FactorSplat uses $(b_L,b_R)=(1,1)$. Both offsets vanish at $\tfbase$, preserving
the shared base appearance. The gate $\gamma_i(\tf)$ applies binary region
removal as defined below. At inference, conditioned SH and opacity are cached
once per TF switch; camera motion then uses the unchanged N-DGS rasterizer.

\noindent\textbf{Analytic visibility gate.}
Authored binary region visibility is applied analytically.  Let
$v_{\tf}(m)\in\{0,1\}$ be the renderer's authored visibility flag for region
$m$, and define the indicator of removal relative to the base preset as
$r_{\tf}(m)=v_{\tfbase}(m)[1-v_{\tf}(m)]$.  The removed material mass of
primitive $i$ is
\begin{equation}
\eta_i(\tf)=\sum_{m,b}p_i(m,b)r_{\tf}(m),\qquad
\gamma_i(\tf)=\mathbf 1[\eta_i(\tf)<\tau],
\quad \tau=\tfrac12 .
\label{eq:visibility-gate}
\end{equation}
Primitives without descriptor support use $\gamma_i=1$.  Equation
\ref{eq:complete-view} multiplies this gate \emph{after} the sigmoid, so a
gated primitive has exactly zero opacity. For mixed support, removal occurs
when the combined mass of removed regions reaches one half. Defining removal
relative to $\tfbase$ preserves the base render. This parameter-free gate
handles binary visibility; the conditional branches handle graded appearance.

\subsection{Initialization and training}
\label{subsec:training}

Conditioning appearance is useful only if the shared geometry represents
structures that later edits reveal. We initialize Gaussians on a preset-neutral
grid covering the union of segmented structures, including those transparent
under $\tfbase$. We optimize
the standard photometric loss through the conditioned
renderer over a mixed-preset batch stream,
\begin{equation}
\widehat{\mathbf I}=\mathcal R(\mathcal G;\pi,\tf),\qquad
\mathcal{L}=
\mathbb{E}_{(\mathbf{I},\pi,\tf)}\Big[
(1-\lambda_{s})\,\mathcal{L}_{1}(\widehat{\mathbf I},\mathbf I)
+\lambda_{s}\big(1-\operatorname{SSIM}(\widehat{\mathbf I},\mathbf I)\big)\Big],
\label{eq:loss}
\end{equation}
where $\mathbf I$ is the reference RGB image for camera $\pi$ and preset
$\tf$, $\mathcal G$ denotes all Gaussian and conditioning parameters,
$\mathcal R$ is the unmodified tile rasterizer, and $\mathcal L_1$ is mean
absolute RGB error. SSIM measures structural similarity~\citep{wang2004ssim};
we use $\lambda_s=0.2$. Gradients reach the factors,
encoder, and base parameters through the same compositing as at test time.
Each training frame supplies its sampled TF. Anchor cameras shared across
presets provide paired appearance changes, while per-preset cameras add view
coverage.

\noindent\textbf{Coverage-aware pruning.}
A primitive needed by one preset may be nearly transparent under another.
We therefore replace the base-opacity pruning score $\sigma(o_i)$ with
$\kappa_i=\max_{\tf\in\mathcal T_{\mathrm{train}}}\bar\alpha_i(\tf)$.
A primitive survives this opacity test if its opacity exceeds the pruning
threshold under any training preset. The score excludes camera-dependent
attenuation $s_i^{\mathrm{view}}$; standard geometric and screen-size pruning
criteria remain active.

\noindent\textbf{Persistent material descriptors.}
The samples $(m_{ik},b_{ik})$ are attached when the preset-neutral grid is
initialized. Their weights measure the initial Gaussian support at each sampled voxel,
\begin{equation}
w_{ik}\;\propto\;
\exp\!\left(-\tfrac12(\mathbf x_{ik}-\bm\mu_i)^{\!\top}
\bm\Sigma_i^{-1}(\mathbf x_{ik}-\bm\mu_i)\right),
\qquad \sum_k w_{ik}=1 ,
\label{eq:density-weights}
\end{equation}
where $(\bm\mu_i,\bm\Sigma_i)$ are the initial center and covariance and
$\mathbf x_{ik}$ is a sampled voxel center. Equation~\ref{eq:density-weights}
makes the descriptor scale-aware: a sub-voxel primitive concentrates on its
own material, whereas a larger primitive integrates over its footprint.

Identifiers and weights remain fixed as each primitive's geometry is
optimized. Clone and split children inherit descriptors and factor rows from
the nearest pre-existing primitive; pruning applies one mask to all attributes.
TF changes therefore use persistent material support rather than reassigning
regions from the optimized centers.

\section{Experiments}
\label{sec:experiments}
\subsection{Protocol}
\noindent\textbf{Data and training.}
We evaluate seven scans: five CT (\texttt{heart}, \texttt{vascular},
\texttt{intestine}, \texttt{lower}, and \texttt{kneejoint}) and two MR
(\texttt{nose} and \texttt{hand}), with 3--17 authored region curves per scan.
Each scan uses 41 conditions: 24 training presets, 4 validation presets,
4 interpolation tests, 4 unseen compositions of individually observed
operations, and 5 OOD tests. Interpolation varies edits within the training
range, while composition combines familiar operations in unseen ways.
OOD includes two functional stress tests (OOD-F: grayscale and opacity
sharpening) and three anatomy-directed edits (OOD-C: target isolation,
occluder suppression, and target emphasis with translucent context). These
test new operations rather than new scans; all models are fitted per scan.
Each training preset has
48 views (12 shared anchors and 36 preset-specific cameras), while all presets
share 40 held-out cameras. We generate ground-truth RGB images at $1600^2$
using cinematic volume path tracing of the original CT or MR volumes and
segmentation masks. Each preset applies region-specific intensity-to-color
and opacity curves. Matched-view renders keep cameras, lighting, and sampling
settings fixed across TFs. All methods use the same held-out cameras and
30{,}000 iterations; conditional models share the 24 training presets.
N-DGS-based methods use volume-derived point initializations. VEG samples
its native random initialization within the source bounds and with the
source point count.
FactorSplat uses rank $r=8$, 64 hidden units, and Mip-Splatting's 3D
filter~\citep{yu2023mipsplattingaliasfree3dgaussian}.

\noindent\textbf{TF-aware metrics.}
\label{subsec:metrics}
We report full-frame PSNR for reconstruction quality and a paired-view metric
for edit fidelity. The latter isolates the predicted appearance change from
the base reconstruction, including edits confined to small regions. Let
$\mathbf D^{\mathrm{gt}}=\mathbf I^{\mathrm{gt}}(\tf)-
\mathbf I^{\mathrm{gt}}(\tfbase)$ and define $\mathbf D^{\mathrm{pred}}$
analogously at the same camera. The changed region is
$M_{\tf}=\{x:\lVert\mathbf D^{\mathrm{gt}}(x)\rVert_\infty>0.04\}$. For
$\Omega\subseteq\mathcal P$, the transfer-function delta error is
\begin{equation}
\deltamet(\tf;\Omega)=\frac{1}{3|\Omega|}\sum_{x\in\Omega}
\left\lVert \mathbf D^{\mathrm{pred}}(x)-
\mathbf D^{\mathrm{gt}}(x)\right\rVert_{1}.
\label{eq:delta}
\end{equation}
Here $\mathcal P$ is the image pixel domain and RGB values lie in $[0,1]$.
We report changed-region error with $\Omega=M_{\tf}$. Unchanged-region leak
instead averages $\lVert\mathbf D^{\mathrm{pred}}(x)\rVert_1$ over
$\mathcal P\setminus M_{\tf}$. For an unconditioned proxy,
$\deltamet(\tf;M_{\tf})$ equals the magnitude of the reference edit and thus
defines the identity floor.

\noindent\textbf{Comparisons.}
The \emph{N-DGS identity floor} uses FactorSplat's pooled training images
without TF input. Each \emph{N-DGS specialist} fits 48 views of one target TF,
including TFs withheld from conditional-model training. FactorSplat and VEG
pool all 24 training presets. Specialists thus receive target-TF supervision
but less camera coverage; they are preset-specific references, not quality
upper bounds. We use the official VEG
implementation~\citep{dyken2026veg}, extending its scalar lookup with a fixed
region per Gaussian to select the corresponding authored curve. VEG retains
its native shading, optimizer, density control, and 100-bin intensity lookup.
In figures and
tables, VEG denotes the floating-point model; its compressed export is labeled
native-quantized. Appendix~\ref{app:veg} specifies the region-aware extension.

\subsection{Reconstruction quality and edit fidelity}

\input{average_comparison_table}

\begin{figure}[!t]
\centering
\includegraphics[width=0.99\textwidth]{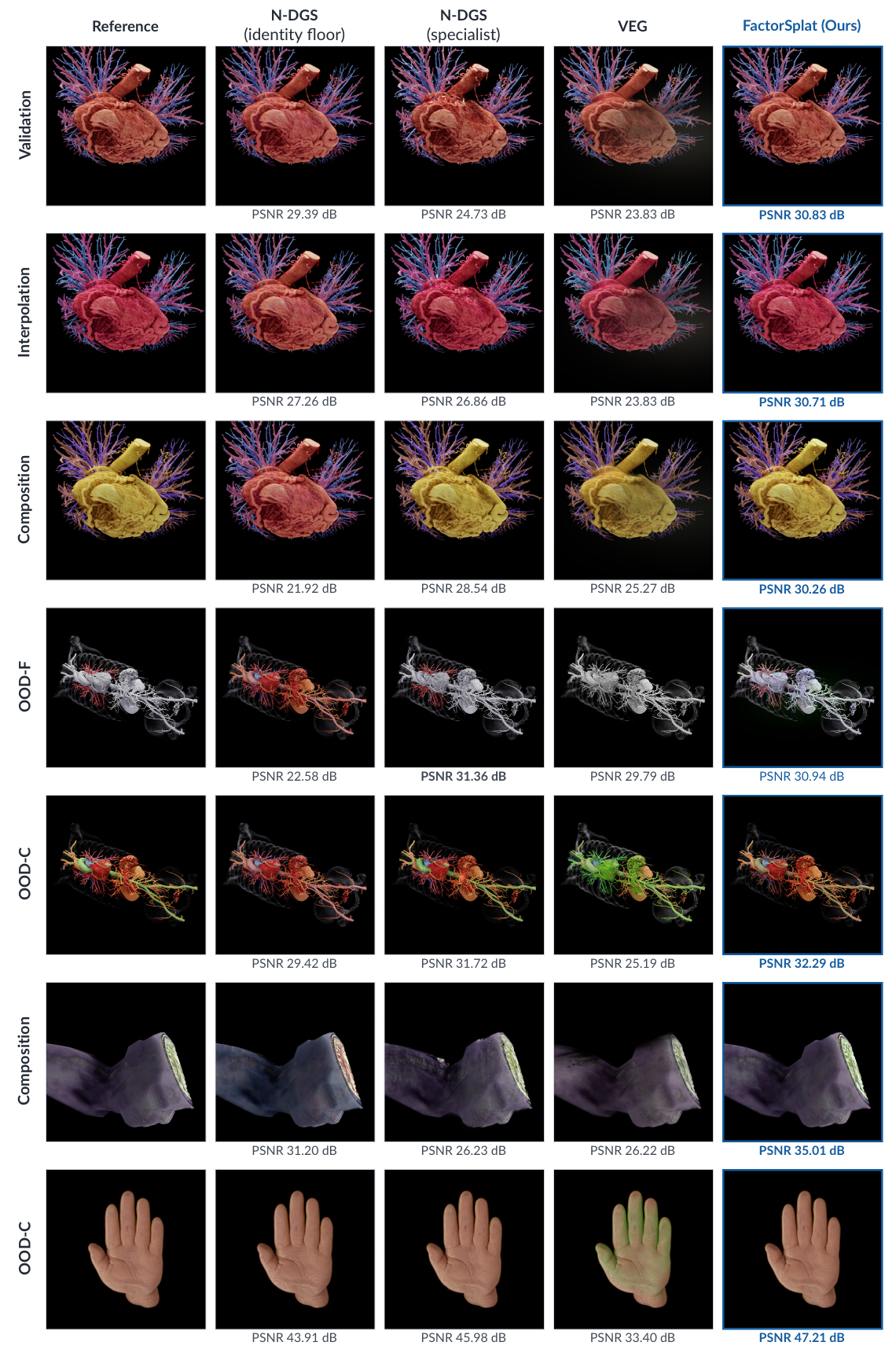}
\caption{\textbf{Held-out CT and MR examples.} Heart (rows 1--3), vascular (4--5),
nose (6), and hand (7). Specialists are trained separately on each target TF.
Full-frame PSNR; bold marks each row's best.}
\label{fig:qualitative-heart}
\vspace{-1em}
\end{figure}

Table~\ref{tab:main-comparison} summarizes the comparison with the N-DGS
references and region-aware VEG as separate CT and MR means, weighting scans
equally within each modality. Appendix Table~\ref{tab:full-comparison} reports
each scan and the seven-scan mean. FactorSplat improves PSNR over
floating-point VEG in 24 of 28 scene--split comparisons, with mean gains of
1.45/1.52/1.10/1.41~dB
on validation, interpolation, composition, and OOD. Changed-region error is
lower in 19 of 28 comparisons, with lower means on every split for both CT
and MR. The two metrics capture distinct aspects of the result: on heart,
FactorSplat has higher PSNR on validation, interpolation, and composition,
while VEG more accurately reproduces the change from the base image.

\noindent\textbf{Shared versus preset-specific models.}
FactorSplat has higher PSNR than specialists on validation and interpolation
for all seven scans, by up to 4.0 and 3.6~dB. On composition it leads on four
scans, ties on one, and trails on two. Its shared geometry receives broader
camera coverage across presets than each 48-view specialist. This may
contribute to the observed advantage, which does not isolate the effect of
conditioning. Specialists retain a 5.2~dB mean advantage on OOD, highlighting
the value of direct target-TF fitting. VEG also leads on heart OOD and several
kneejoint splits; new edit operations remain harder than interpolation.

Figure~\ref{fig:qualitative-heart} shows matched held-out CT and MR views,
including differences in tissue color and visibility under unseen edits.
Figure~\ref{fig:qualitative-vascular} provides additional examples.

\subsection{Ablation studies}
Table~\ref{tab:ablation} isolates the physical lookup, learned residual,
functional input, and visibility gate on heart and vascular. Appendix
Table~\ref{tab:ablation-delta} reports the corresponding edit errors.

\begin{table}[t]
\centering
\caption{\textbf{Two-scene ablations ($r{=}8$).} PSNR$\uparrow$, mean rank,
and worst gap over eight comparisons.}
\label{tab:ablation}
\setlength{\tabcolsep}{3pt}
\resizebox{\textwidth}{!}{%
\begin{tabular}{@{}l cccc c cccc c cc@{}}
\toprule
 & \multicolumn{4}{c}{\textbf{\texttt{heart}}} &&
   \multicolumn{4}{c}{\textbf{\texttt{vascular}}} &&
   \multicolumn{2}{c}{\textbf{robustness}} \\
\cmidrule{2-5}\cmidrule{7-10}\cmidrule{12-13}
Variant & val & int. & cmp & OOD
        && val & int. & cmp & OOD
        && rank$\downarrow$ & gap (dB)$\downarrow$ \\
\midrule
\factorsplat{} (Full)
 & \textbf{30.79} & \textbf{29.33} & \textbf{28.68} & \textbf{23.77}
 && 34.20 & \textbf{33.68} & 32.59 & \textbf{30.88}
 && \textbf{1.69} & \textbf{0.48} \\
$-$ exact gate
 & \textbf{30.79} & \textbf{29.33} & \textbf{28.68} & 23.45
 && 34.20 & \textbf{33.68} & 32.59 & 30.36
 && 1.94 & 0.52 \\
$-$ learned residual
 & 30.16 & 28.69 & 27.61 & 23.22
 && 33.70 & 32.55 & 31.92 & 29.28
 && 3.62 & 1.60 \\
$-$ physical lookup
 & 29.77 & 27.48 & 26.11 & 20.30
 && 33.31 & 32.13 & 29.94 & 29.18
 && 4.88 & 3.47 \\
$-$ functional code
 & 30.76 & 28.38 & 28.42 & 22.75
 && \textbf{34.30} & \textbf{33.68} & \textbf{33.07} & 28.04
 && 2.88 & 2.84 \\
\bottomrule
\end{tabular}%
}
\vspace{-1em}
\end{table}

\noindent\textbf{Complementary appearance branches.}
Removing the learned residual reduces PSNR in all eight columns; removing
the physical lookup causes larger losses, including 2.57 and 2.65~dB on
composition. These results support combining a direct response to the authored
edit with learned corrections to reproduce the reference images.

\noindent\textbf{Functional input.}
Replacing the functional code with preset embeddings assigns an unseen TF
the embedding of its nearest training preset. This variant is competitive on nearby vascular edits, but loses 1.02 and
2.84~dB on heart and vascular OOD. The functional input is most useful when
the requested curve differs substantially from the training presets.

\noindent\textbf{Visibility control.}
Removing the analytic gate leaves validation, interpolation, and composition
unchanged, while reducing OOD PSNR by 0.32 and 0.52~dB. OOD delta error rises
from 9.64 to 10.87 on heart and from 4.54 to 5.01 on vascular
($\times10^{-2}$). The gate benefits material removal.

\subsection{Computational cost}
\label{sec:computational-cost}
\noindent\textbf{Rendering.}
FactorSplat renders faster than VEG, averaging 524 FPS versus 144 FPS across
seven scans on B200 GPUs at $1600^2$ (Table~\ref{tab:main-comparison}).
Its cached fast renderer separates appearance updates from view navigation:
conditioning is evaluated once per TF switch, averaging 1.17~ms; cached
attributes are reused during camera motion. Both methods support new
TFs without refitting. For FactorSplat, FPS uses the fast renderer and PSNR
the quality renderer; Appendix~\ref{app:efficiency} details timing.

\noindent\textbf{Training.}
FactorSplat also trains faster than VEG: mean training time is 9.8 versus
43.8 minutes per scan for 30k iterations on a B200. One
FactorSplat checkpoint supports multiple appearances, whereas N-DGS
specialists must be trained separately for every target TF. Across the
17 evaluated TFs, their estimated total training cost averages 88.1 minutes
per scan, computed as $17\times$ one fixed preset's training time.
Sharing geometry and conditioning appearance therefore avoids the repeated
optimization required by per-TF specialists.

\FloatBarrier
\section{Discussion}
\label{sec:discussion}

\noindent\textbf{Implications for TF authoring.}
Function-valued conditioning lets users adjust region-specific color and
opacity without selecting a different checkpoint. Interpolation and
composition probe two complementary aspects of this workflow: adjusting
familiar controls and combining edits observed separately during training.
The results support both forms of exploration with a shared proxy. They
also suggest a useful principle for training-bank design: include distinct
appearance operations and expose the materials they affect, rather than
simply adding visually similar presets. TF diversity supplies appearance
supervision, while coverage-aware pruning retains primitives needed to
express those changes. A training bank that combines both types of coverage
could extend the range of edits supported by one representation.

\noindent\textbf{Local control and global appearance.}
Local conditioning ties each primitive's response to the materials it
represents, making authored edits composable within a shared Gaussian field.
This locality also limits the response to global illumination changes:
edits outside a primitive's support leave its code unchanged, even when
they alter its lighting. Contextual conditioning could address this limitation.
Persistent descriptors can also misalign as centers move, and visibility
gating removes whole Gaussians rather than clipping at voxel boundaries.

\noindent\textbf{Feed-forward proxy generation.}
\factorsplat{} learns one proxy per scan. Feed-forward prediction of shared
geometry and conditional appearance factors, building on
Render-FM~\citep{gao2025renderfm}, could reduce this construction cost while
retaining inference-time appearance control.

\section{Conclusion}
\label{sec:conclusion}
FactorSplat combines local TF lookup and a low-rank appearance residual to
render unseen region-specific transfer functions with shared Gaussian geometry.
Across seven CT and MR scans, it improves mean reconstruction and edit fidelity
over region-aware VEG while reducing training time and increasing rendering
throughput. The ablations support combining authored optical changes with
learned corrections. By decoupling appearance updates from camera motion,
FactorSplat supports interactive color and opacity control within one proxy
instead of separate models for each preset. Generalizing to a broader range
of unseen edit operations remains an important next step.

\clearpage
\subsection*{AI use statement}
Generative AI tools assisted with method development, experimental design,
implementation, result interpretation, and manuscript revision. The authors made the final research
decisions and take responsibility for the content.

\subsection*{Reproducibility statement}
Sections~\ref{sec:method} and~\ref{sec:experiments} specify the model, training
procedure, TF and camera splits, baselines, and metrics. The appendix provides
the cached evaluation procedure, VEG adaptation, complete per-scan results,
ablation edit metrics, and timing protocol. The repository retains experiment
scripts, TF manifests, per-view metrics, and figure-source records.

\subsection*{Ethics statement}
FactorSplat is a research visualization method evaluated on rendered CT and
MR images, not a clinically validated diagnostic system. Approximation errors
and TF edits can change anatomical visibility. Clinical use would require
task-specific validation, expert oversight, and access to the original scan.
Medical volumes, masks, and derived artifacts remain subject to source
permissions, privacy requirements, and applicable licenses.

\clearpage
\bibliographystyle{iclr2027_conference}
\bibliography{refs}

\clearpage
\appendix

\section{Efficient transfer-function evaluation}
\label{app:implementation}
\noindent\textbf{Appearance scales.}
All scans use lookup scales $\lambda_L^c=1$ and $\lambda_L^\alpha=4$,
and residual scales $\lambda_R^c=0.25$ and $\lambda_R^\alpha=4$.
The lookup gains are learned, while these output scales remain fixed.

\noindent\textbf{Cached transfer-function evaluation.}
At each TF switch, the shared encoder evaluates the $M\times B$ region--intensity
entries once, where $M$ is the number of region-specific transfer curves and
$B=64$ is the number of intensity bins per curve. Each Gaussian aggregates the resulting $r$-dimensional
embeddings over its persistent local support to obtain $\zitf$. The physical
lookup and low-rank residual then update its DC color and opacity. These
attributes are cached across camera changes, avoiding repeated TF encoding
during view navigation.

\noindent\textbf{Representation size.}
The conditional model adds $4Nr$ per-Gaussian factors, $H(4+r)$ shared encoder
weights, and four lookup gains. Its persistent descriptors contain at most
$NK$ weighted region--intensity samples. The representation therefore scales
linearly with the number of primitives, independently of the number of TFs.

\section{Region-aware VEG baseline}
\label{app:veg}
We build on the official VEG code~\citep{dyken2026veg}. To support the
multi-region condition $\mathbf R_{\tf}(m,h)$, our extension assigns each
Gaussian a fixed categorical region $m_i$ and one learnable scalar $v_i$.
We retain the released initializer: the input point cloud supplies its
bounding box and point count, while positions, scalars, and normals follow
VEG's native random draws. Each primitive receives its nearest source seed's
density-weighted majority region. The authored curves are resampled to
VEG's native $B_{\mathrm V}=100$ intensity knots, giving $\mathbf R_{\tf}^{(B_{\mathrm V})}$.
Its within-region coordinate is
$h_i=(B_{\mathrm V}-1)\operatorname{sigmoid}(v_i)$. Linear interpolation is performed
only between adjacent intensity bins of the same region:
\begin{equation}
  (\mathbf c_i(\tf),a_i(\tf))=
  \operatorname{lerp}\!\left(\mathbf R_{\tf}^{(B_{\mathrm V})}(m_i,\lfloor h_i\rfloor),
  \mathbf R_{\tf}^{(B_{\mathrm V})}(m_i,\lceil h_i\rceil),h_i-\lfloor h_i\rfloor\right).
\end{equation}
The queried color enters VEG's Blinn--Phong shading with learned normals,
ambient, diffuse, and specular factors, and shininess. Opacity is
$w_i a_i(\tf)$, where $w_i$ is VEG's independent sigmoid-activated weight.
The authored region-visibility flag is included in TF alpha. Region labels
remain fixed under motion and follow the actual parent indices during native
cloning and splitting. We retain the upstream optimizer, image losses,
regularizers, and weight-based pruning, with scalar and opacity-weight
learning rates $2.5\times10^{-4}$ and $2.5\times10^{-2}$, respectively.
Training uses 24 authored presets for 30{,}000 iterations in place of VEG's
synthetic opacity bank, with no primitive-count cap.

\noindent\textbf{Floating-point and native-quantized evaluation.}
Both rows use the same trained checkpoint. The float row evaluates its
uncompressed parameters. The native-quantized row follows the released
deployment path: 256-center codebooks encode opacity, scale, rotation,
normal, scalar, diffuse, and ambient attributes; codebook centers and the
remaining continuous attributes use half precision. We evaluate decoded
parameters, with region assignments preserved losslessly. Quality and timing
are measured separately for each representation.
Quantization lowers mean PSNR by 0.15/0.12/0.10/0.05~dB on
validation/interpolation/composition/OOD.

\noindent\textbf{Numerical stability.}
We fix a numerical bug in the released VEG implementation that produces
non-finite gradients and interrupts training. The native specular term
$|\mathbf n_i^\top\mathbf h|^{s_i}$ has a
singular derivative at zero when $0<s_i<1$, where $\mathbf h$ is the
half-vector and $s_i$ is shininess. We mask exact-zero inputs before
exponentiation and assign zero gradients at those inputs, preserving forward
values and gradients elsewhere. With this correction, VEG completes training
on all seven scans.

\section{Per-scan results}
\label{app:per-scan}
Table~\ref{tab:full-comparison} reports each scan separately and the
unweighted seven-scan mean. Each split averages its held-out TFs, with five
edits contributing to OOD. FactorSplat leads VEG in PSNR across all splits
on vascular, intestine, lower, nose, and hand; VEG has lower edit error
throughout heart and kneejoint.
\input{full_comparison_table}
\FloatBarrier

\section{Ablation edit fidelity}
\label{app:ablation-delta}
Table~\ref{tab:ablation-delta} complements the reconstruction metrics in
Table~\ref{tab:ablation} with changed-region $\deltamet$ for the same variants.
Removing either appearance branch increases edit error in all eight columns.
Preset embeddings remain competitive on nearby vascular conditions, whereas
the functional code improves OOD edit fidelity on both scans. The visibility
gate affects only presets that explicitly remove material.
Unchanged-region leak spans $0.12$--$0.43\times10^{-2}$ across these variants.
In Table~\ref{tab:ablation}, mean rank averages ranks across eight PSNR columns;
worst gap is the maximum deficit from a column's best score.

\begin{table}[!htbp]
\centering
\caption{\textbf{Ablation edit fidelity.} Changed-region $\deltamet$
($\times10^{-2}$; lower is better).}
\label{tab:ablation-delta}
\setlength{\tabcolsep}{3.6pt}
\begin{tabular}{@{}l cccc c cccc@{}}
\toprule
 & \multicolumn{4}{c}{\textbf{\texttt{heart}}} &&
   \multicolumn{4}{c}{\textbf{\texttt{vascular}}} \\
\cmidrule{2-5}\cmidrule{7-10}
Variant & val & int. & cmp & OOD
        && val & int. & cmp & OOD \\
\midrule
\factorsplat{} (Full)
 & \textbf{1.31} & \textbf{2.63} & \textbf{3.14} & \textbf{9.64}
 && 1.60 & 2.48 & 3.59 & \textbf{4.54} \\
$-$ exact gate
 & \textbf{1.31} & \textbf{2.63} & \textbf{3.14} & 10.87
 && 1.60 & 2.48 & 3.59 & 5.01 \\
$-$ learned residual
 & 1.39 & 2.87 & 3.62 & 10.56
 && 1.67 & 3.40 & 4.26 & 6.83 \\
$-$ physical lookup
 & 2.86 & 5.27 & 6.59 & 14.71
 && 2.95 & 4.67 & 6.98 & 7.77 \\
$-$ functional code
 & 1.35 & 3.46 & 3.17 & 10.92
 && \textbf{1.50} & \textbf{2.43} & \textbf{3.17} & 7.85 \\
\bottomrule
\end{tabular}
\end{table}

\FloatBarrier

\section{Runtime measurements}
\label{app:efficiency}

\noindent\textbf{Rendering protocol.}
Table~\ref{tab:matched-efficiency} compares the two methods across seven scans on
B200 GPUs at 1965~MHz. Each method renders the same eight validation views
with five warm-ups and 20 CUDA-event measurements per view. FPS is the inverse
of mean frame time; Table~\ref{tab:main-comparison} averages it equally across
scans within each modality. N-DGS references use the same fast rasterizer and
timing protocol, with the appropriate specialist selected before timing.
Measurements include VEG's native TF lookup, shading, rasterization, and
auxiliary outputs, and FactorSplat's cached fast rendering; data loading,
image saving, and scoring are excluded. PSNR is measured separately with
the quality renderer over all 680 held-out views (17 TFs at 40 cameras each).
Fast and quality rendering can produce different images, so the table
compares deployment throughput alongside reconstruction quality.
Primitive counts provide context for the rendering cost.

\input{matched_efficiency_table}

\noindent\textbf{Transfer-function updates.}
TF-update latency measures appearance-cache construction separately, with
five warm-ups and 20 synchronized wall-time measurements per view. This cost
is incurred once per TF switch, not during fixed-TF view navigation.

\clearpage
\section{Additional qualitative comparisons}
\label{app:qualitative}

\noindent\begin{minipage}{\textwidth}
\centering
\includegraphics[width=0.94\textwidth]{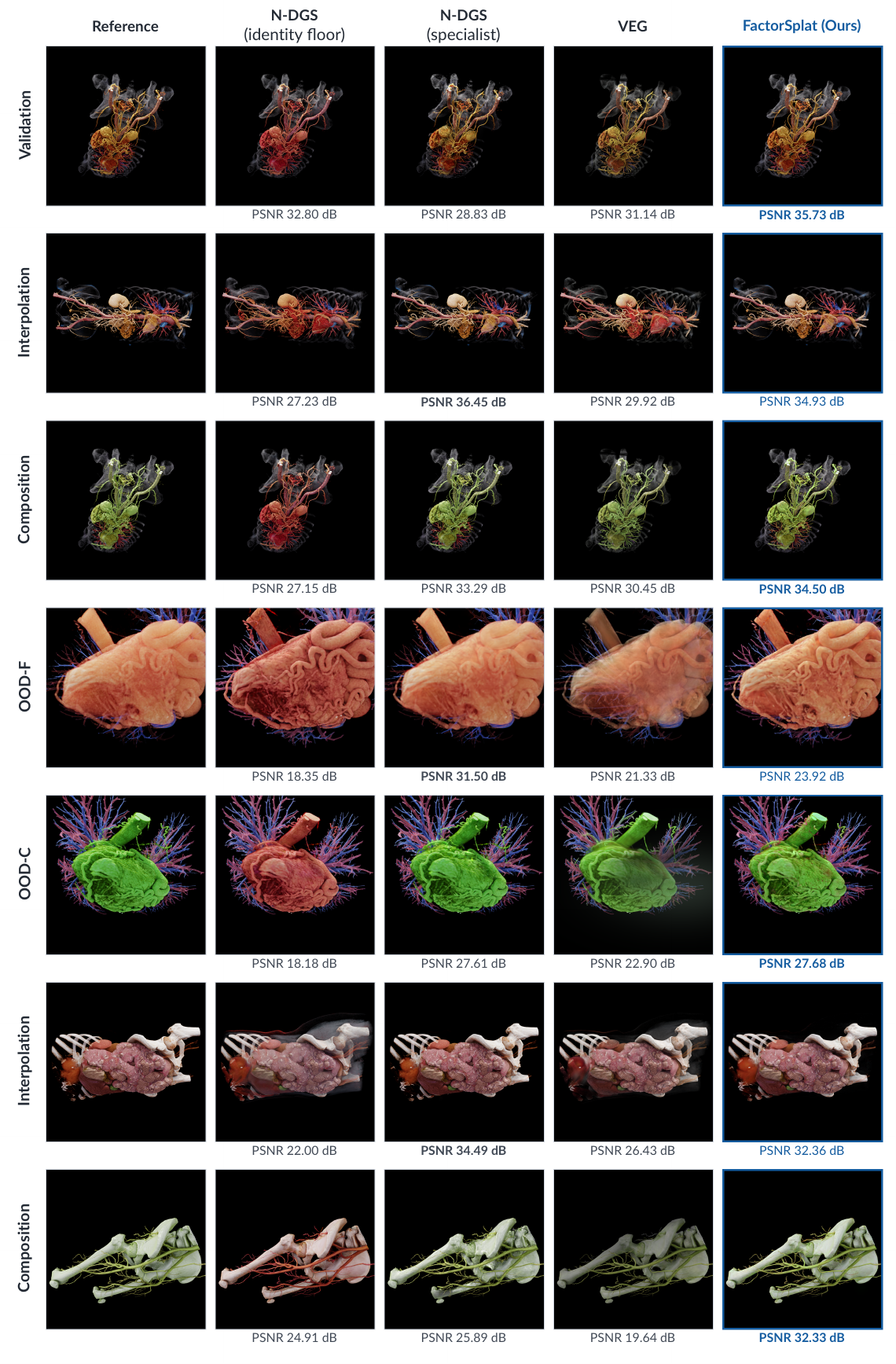}
\captionof{figure}{\textbf{Additional held-out CT examples.} Vascular (rows 1--3), heart
(4--5), intestine (6), and lower body (7). Full-frame PSNR; bold marks each row's best.}
\label{fig:qualitative-vascular}
\end{minipage}

\end{document}

%% file: math_commands.tex
\usepackage{amsmath,amsfonts,bm}

\def\eqref#1{equation~\ref{#1}}

\def\1{\bm{1}}

\DeclareMathAlphabet{\mathsfit}{\encodingdefault}{\sfdefault}{m}{sl}
\SetMathAlphabet{\mathsfit}{bold}{\encodingdefault}{\sfdefault}{bx}{n}

%% file: average_comparison_table.tex
\begin{table}[t]
\centering
\caption{\textbf{CT/MR quality and efficiency.} Identity floor: no TF input; specialists: per-target-TF training. B200 FPS at $1600^2$; OOD: five edits. Per-scan results: Table~\ref{tab:full-comparison}; timing: Section~\ref{sec:computational-cost}.}
\label{tab:main-comparison}
\small
\setlength{\tabcolsep}{2.2pt}
\renewcommand{\arraystretch}{1.25}
\begin{tabular}{@{}c l cccc c cccc @{\hspace{7pt}} r r@{}}
\toprule
 & & \multicolumn{4}{c}{$\deltamet\!\downarrow$ ($\times10^{-2}$)} &&
   \multicolumn{4}{c}{PSNR$\uparrow$} & \multirow{2}{*}{FPS$\uparrow$} & \multirow{2}{*}{\shortstack{Train\\(min)$\downarrow$}} \\
\cmidrule{3-6}\cmidrule(r{6pt}){8-11}
 & Model & val & interp & comp & OOD
      && val & interp & comp & OOD & & \\
\midrule
\multirow{5}{*}{\rotatebox{90}{\fontsize{8}{9}\selectfont\texttt{CT (5 scans)}}} & N-DGS (identity floor) & 3.60 & 6.49 & 8.96 & 17.74 && 31.11 & 27.88 & 25.50 & 21.11 & 851 & 6.9 \\
 & N-DGS (specialist) & \multicolumn{4}{c}{---} && 30.70 & 30.82 & 31.09 & \textbf{31.94} & 866 & 92.8 \\
\cdashline{2-13}
\addlinespace[1pt]
 & VEG & 1.40 & 2.55 & 3.05 & 9.15 && 31.80 & 30.90 & 30.64 & 25.37 & 134 & 44.3 \\
 & VEG (native-quantized) & 1.41 & 2.56 & 3.06 & 9.17 && 31.65 & 30.78 & 30.54 & 25.32 & 136 & 44.3 \\
 & \textbf{\factorsplat{} (Ours)} & \textbf{1.36} & \textbf{2.17} & \textbf{2.88} & \textbf{8.09} && \textbf{33.08} & \textbf{32.29} & \textbf{31.33} & 26.58 & 508 & 10.5 \\
\midrule
\multirow{5}{*}{\rotatebox{90}{\fontsize{8}{9}\selectfont\texttt{MR (2 scans)}}} & N-DGS (identity floor) & 2.88 & 4.25 & 5.48 & 11.09 && 36.58 & 35.54 & 31.17 & 27.86 & 925 & 6.0 \\
 & N-DGS (specialist) & \multicolumn{4}{c}{---} && 37.03 & 37.04 & 37.07 & \textbf{38.00} & 930 & 76.4 \\
\cdashline{2-13}
\addlinespace[1pt]
 & VEG & 1.19 & 1.89 & 1.59 & 5.16 && 38.82 & 38.52 & 37.82 & 31.24 & 169 & 42.6 \\
 & VEG (native-quantized) & 1.19 & 1.90 & 1.59 & 5.15 && 38.67 & 38.39 & 37.70 & 31.20 & 171 & 42.6 \\
 & \textbf{\factorsplat{} (Ours)} & \textbf{1.09} & \textbf{1.62} & \textbf{1.28} & \textbf{4.65} && \textbf{40.68} & \textbf{40.36} & \textbf{39.95} & 33.13 & 562 & 8.3 \\
\bottomrule
\end{tabular}
\vspace{-1em}
\end{table}

%% file: full_comparison_table.tex
\begin{table}[t]
\centering
\caption{\textbf{Per-scan comparison.} Changed-region $\deltamet$
($\times10^{-2}$) and PSNR; OOD averages five edits and (h) is the unweighted
mean. Bold marks column best.}
\label{tab:full-comparison}
\small
\setlength{\tabcolsep}{3pt}
\begin{tabular}{l cccc c cccc}
\toprule
 & \multicolumn{4}{c}{$\deltamet\!\downarrow$ ($\times10^{-2}$)} &&
   \multicolumn{4}{c}{PSNR$\uparrow$} \\
\cmidrule{2-5}\cmidrule{7-10}
Model & val & interp & comp & OOD
      && val & interp & comp & OOD \\
\midrule
\multicolumn{10}{l}{\textbf{(a) \texttt{heart}}\;\footnotesize(CT, 12 labels)} \\
N-DGS (identity floor) & 3.74 & 6.76 & 10.28 & 19.92 && 28.31 & 25.18 & 21.49 & 17.77 \\
N-DGS (specialist) & \multicolumn{4}{c}{---} && 28.82 & 29.06 & \textbf{29.21} & \textbf{30.47} \\
\cdashline{1-10}
\addlinespace[1pt]
VEG & \textbf{1.22} & \textbf{2.26} & \textbf{3.03} & \textbf{5.93} && 28.79 & 28.21 & 27.80 & 25.70 \\
VEG (native-quantized) & \textbf{1.22} & \textbf{2.26} & \textbf{3.03} & 5.95 && 28.69 & 28.13 & 27.73 & 25.66 \\
\factorsplat{} (Ours) & 1.31 & 2.63 & 3.14 & 9.64 && \textbf{30.79} & \textbf{29.33} & 28.68 & 23.77 \\
\addlinespace[2pt]
\midrule
\multicolumn{10}{l}{\textbf{(b) \texttt{vascular}}\;\footnotesize(CT, 15 labels)} \\
N-DGS (identity floor) & 3.93 & 7.53 & 10.63 & 13.81 && 32.40 & 29.36 & 26.83 & 25.15 \\
N-DGS (specialist) & \multicolumn{4}{c}{---} && 32.27 & 32.45 & \textbf{32.59} & \textbf{33.16} \\
\cdashline{1-10}
\addlinespace[1pt]
VEG & 1.61 & 3.58 & 4.37 & 10.02 && 31.95 & 30.91 & 30.31 & 25.68 \\
VEG (native-quantized) & 1.62 & 3.59 & 4.39 & 10.03 && 31.47 & 30.55 & 30.03 & 25.56 \\
\factorsplat{} (Ours) & \textbf{1.60} & \textbf{2.48} & \textbf{3.59} & \textbf{4.54} && \textbf{34.20} & \textbf{33.68} & \textbf{32.59} & 30.88 \\
\addlinespace[2pt]
\midrule
\multicolumn{10}{l}{\textbf{(c) \texttt{intestine}}\;\footnotesize(CT, 17 labels)} \\
N-DGS (identity floor) & 3.40 & 6.61 & 8.34 & 14.97 && 30.79 & 26.92 & 25.63 & 21.58 \\
N-DGS (specialist) & \multicolumn{4}{c}{---} && 29.27 & 29.06 & 29.35 & \textbf{30.07} \\
\cdashline{1-10}
\addlinespace[1pt]
VEG & 1.73 & 3.06 & 3.35 & 10.25 && 31.61 & 29.81 & 29.75 & 24.20 \\
VEG (native-quantized) & 1.74 & 3.08 & 3.36 & 10.24 && 31.56 & 29.76 & 29.69 & 24.18 \\
\factorsplat{} (Ours) & \textbf{1.30} & \textbf{2.08} & \textbf{2.85} & \textbf{9.95} && \textbf{32.53} & \textbf{31.60} & \textbf{30.47} & 25.53 \\
\addlinespace[2pt]
\midrule
\multicolumn{10}{l}{\textbf{(d) \texttt{lower}}\;\footnotesize(CT, 7 labels)} \\
N-DGS (identity floor) & 3.69 & 6.95 & 10.10 & 17.33 && 32.68 & 29.07 & 27.00 & 24.10 \\
N-DGS (specialist) & \multicolumn{4}{c}{---} && 32.10 & 32.38 & \textbf{32.86} & \textbf{33.66} \\
\cdashline{1-10}
\addlinespace[1pt]
VEG & 1.33 & 2.42 & \textbf{3.03} & 8.26 && 32.93 & 32.38 & 32.45 & 28.40 \\
VEG (native-quantized) & 1.33 & 2.42 & 3.04 & 8.27 && 32.88 & 32.34 & 32.41 & 28.38 \\
\factorsplat{} (Ours) & \textbf{1.29} & \textbf{2.16} & 3.08 & \textbf{4.61} && \textbf{34.30} & \textbf{33.55} & 32.66 & 30.87 \\
\addlinespace[2pt]
\midrule
\multicolumn{10}{l}{\textbf{(e) \texttt{kneejoint}}\;\footnotesize(CT, 3 labels)} \\
N-DGS (identity floor) & 3.24 & 4.58 & 5.47 & 22.67 && 31.37 & 28.87 & 26.54 & 16.95 \\
N-DGS (specialist) & \multicolumn{4}{c}{---} && 31.05 & 31.13 & 31.44 & \textbf{32.36} \\
\cdashline{1-10}
\addlinespace[1pt]
VEG & \textbf{1.14} & \textbf{1.42} & \textbf{1.47} & \textbf{11.30} && \textbf{33.72} & 33.21 & \textbf{32.91} & 22.85 \\
VEG (native-quantized) & \textbf{1.14} & 1.43 & 1.48 & 11.36 && 33.63 & 33.12 & 32.85 & 22.80 \\
\factorsplat{} (Ours) & 1.31 & 1.47 & 1.73 & 11.74 && 33.59 & \textbf{33.29} & 32.24 & 21.87 \\
\addlinespace[2pt]
\midrule
\multicolumn{10}{l}{\textbf{(f) \texttt{nose}}\;\footnotesize(MR, 3 labels)} \\
N-DGS (identity floor) & 2.96 & 4.14 & 4.48 & 14.38 && 36.60 & 35.02 & 33.04 & 25.06 \\
N-DGS (specialist) & \multicolumn{4}{c}{---} && 34.94 & 34.80 & 34.91 & \textbf{36.85} \\
\cdashline{1-10}
\addlinespace[1pt]
VEG & 1.48 & 2.08 & 1.98 & 6.14 && 36.69 & 36.19 & 35.54 & 29.35 \\
VEG (native-quantized) & 1.48 & 2.08 & 1.98 & 6.13 && 36.51 & 36.02 & 35.40 & 29.30 \\
\factorsplat{} (Ours) & \textbf{1.33} & \textbf{1.71} & \textbf{1.60} & \textbf{5.50} && \textbf{38.27} & \textbf{37.86} & \textbf{37.49} & 30.94 \\
\addlinespace[2pt]
\midrule
\multicolumn{10}{l}{\textbf{(g) \texttt{hand}}\;\footnotesize(MR, 9 labels)} \\
N-DGS (identity floor) & 2.80 & 4.35 & 6.48 & 7.80 && 36.55 & 36.06 & 29.30 & 30.65 \\
N-DGS (specialist) & \multicolumn{4}{c}{---} && 39.12 & 39.29 & 39.23 & \textbf{39.16} \\
\cdashline{1-10}
\addlinespace[1pt]
VEG & 0.90 & 1.71 & 1.20 & 4.18 && 40.94 & 40.85 & 40.09 & 33.13 \\
VEG (native-quantized) & 0.90 & 1.71 & 1.20 & 4.18 && 40.83 & 40.75 & 40.00 & 33.11 \\
\factorsplat{} (Ours) & \textbf{0.85} & \textbf{1.53} & \textbf{0.96} & \textbf{3.79} && \textbf{43.09} & \textbf{42.87} & \textbf{42.40} & 35.32 \\
\addlinespace[2pt]
\midrule
\multicolumn{10}{l}{\textbf{(h) average}} \\
N-DGS (identity floor) & 3.39 & 5.85 & 7.97 & 15.84 && 32.67 & 30.07 & 27.12 & 23.04 \\
N-DGS (specialist) & \multicolumn{4}{c}{---} && 32.51 & 32.59 & 32.80 & \textbf{33.67} \\
\cdashline{1-10}
\addlinespace[1pt]
VEG & 1.34 & 2.36 & 2.63 & 8.01 && 33.80 & 33.08 & 32.69 & 27.05 \\
VEG (native-quantized) & 1.35 & 2.37 & 2.64 & 8.02 && 33.65 & 32.95 & 32.59 & 27.00 \\
\factorsplat{} (Ours) & \textbf{1.29} & \textbf{2.01} & \textbf{2.42} & \textbf{7.11} && \textbf{35.25} & \textbf{34.60} & \textbf{33.79} & 28.45 \\
\bottomrule
\end{tabular}
\end{table}

%% file: matched_efficiency_table.tex
\begin{table}[!htbp]
\centering
\caption{\textbf{B200 deployment at $1600^2$.} Matched views; FactorSplat FPS uses cached fast rendering. PSNR uses quality rendering.}
\label{tab:matched-efficiency}
\setlength{\tabcolsep}{4pt}
\begin{tabular}{@{}l rr rr rr r@{}}
\toprule
 & \multicolumn{2}{c}{$N$ (k)} & \multicolumn{2}{c}{PSNR$\uparrow$} & \multicolumn{2}{c}{FPS$\uparrow$} & \\
Scene & Ours & VEG & Ours & VEG & Ours & VEG & Speedup \\
\midrule
\texttt{heart} & 278.6 & 146.3 & 27.89 & 27.51 & 477 & 109 & 4.36$\times$ \\
\texttt{vascular} & 235.7 & 148.1 & 32.72 & 29.47 & 525 & 181 & 2.90$\times$ \\
\texttt{intestine} & 281.8 & 208.9 & 29.77 & 28.57 & 494 & 122 & 4.06$\times$ \\
\texttt{lower} & 250.5 & 127.0 & 32.73 & 31.36 & 531 & 155 & 3.42$\times$ \\
\texttt{kneejoint} & 80.4 & 82.7 & 29.75 & 30.21 & 515 & 103 & 5.03$\times$ \\
\texttt{nose} & 150.2 & 111.9 & 35.84 & 34.14 & 564 & 146 & 3.87$\times$ \\
\texttt{hand} & 135.1 & 54.7 & 40.59 & 38.42 & 560 & 193 & 2.89$\times$ \\
\bottomrule
\end{tabular}
\end{table}

%% file: main.bbl
\begin{thebibliography}{25}
\providecommand{\natexlab}[1]{#1}
\providecommand{\url}[1]{\texttt{#1}}
\expandafter\ifx\csname urlstyle\endcsname\relax
  \providecommand{\doi}[1]{doi: #1}\else
  \providecommand{\doi}{doi: \begingroup \urlstyle{rm}\Url}\fi

\bibitem[Berger et~al.(2019)Berger, Li, and Levine]{berger2019generative}
Matthew Berger, Jixian Li, and Joshua~A. Levine.
\newblock A generative model for volume rendering.
\newblock \emph{IEEE Transactions on Visualization and Computer Graphics},
  25\penalty0 (4):\penalty0 1636--1650, 2019.
\newblock \doi{10.1109/TVCG.2018.2816059}.

\bibitem[Dappa et~al.(2016)Dappa, Higashigaito, Fornaro, Leschka, Wildermuth,
  and Alkadhi]{dappa2016cinematic}
Evelyn Dappa, Kai Higashigaito, J{\"u}rgen Fornaro, Sebastian Leschka, Simon
  Wildermuth, and Hatem Alkadhi.
\newblock Cinematic rendering---an alternative to volume rendering for {3D}
  computed tomography imaging.
\newblock \emph{Insights into Imaging}, 7\penalty0 (6):\penalty0 849--856,
  2016.
\newblock \doi{10.1007/s13244-016-0518-1}.

\bibitem[Dyken et~al.(2026)Dyken, Sewell, Usher, Debardeleben, Petruzza, and
  Kumar]{dyken2026veg}
Landon Dyken, Andres Sewell, Will Usher, Nathan Debardeleben, Steve Petruzza,
  and Sidharth Kumar.
\newblock Volume encoding gaussians: Transfer function-agnostic {3D} gaussians
  for volume rendering.
\newblock \emph{IEEE Transactions on Visualization and Computer Graphics},
  32\penalty0 (6):\penalty0 4883--4893, 2026.
\newblock \doi{10.1109/TVCG.2026.3694460}.

\bibitem[Engel et~al.(2024)Engel, Sick, and Ropinski]{engel2024segmentationtf}
Dominik Engel, Leon Sick, and Timo Ropinski.
\newblock Leveraging self-supervised vision transformers for segmentation-based
  transfer function design.
\newblock \emph{IEEE Transactions on Visualization and Computer Graphics},
  2024.
\newblock \doi{10.1109/TVCG.2024.3401755}.

\bibitem[Gao et~al.(2025{\natexlab{a}})Gao, Planche, Zheng, Choudhuri, Chen,
  and Wu]{gao20246dgs}
Zhongpai Gao, Benjamin Planche, Meng Zheng, Anwesa Choudhuri, Terrence Chen,
  and Ziyan Wu.
\newblock 6dgs: Enhanced direction-aware gaussian splatting for volumetric
  rendering.
\newblock In \emph{ICLR}, 2025{\natexlab{a}}.

\bibitem[Gao et~al.(2025{\natexlab{b}})Gao, Planche, Zheng, Choudhuri, Chen,
  and Wu]{gao20257dgs}
Zhongpai Gao, Benjamin Planche, Meng Zheng, Anwesa Choudhuri, Terrence Chen,
  and Ziyan Wu.
\newblock 7dgs: Unified spatial-temporal-angular gaussian splatting.
\newblock In \emph{ICCV}, 2025{\natexlab{b}}.

\bibitem[Gao et~al.(2026)Gao, Planche, Zheng, Choudhuri, Nguyen, Chen, and
  Wu]{gao2025renderfm}
Zhongpai Gao, Benjamin Planche, Meng Zheng, Anwesa Choudhuri, Van~Nguyen
  Nguyen, Terrence Chen, and Ziyan Wu.
\newblock Render-{FM}: Feedforward model for real-time photorealistic
  volumetric rendering.
\newblock In \emph{ECCV}, 2026.
\newblock arXiv:2505.17338.

\bibitem[Ha et~al.(2017)Ha, Dai, and Le]{ha2017hypernetworks}
David Ha, Andrew~M. Dai, and Quoc~V. Le.
\newblock Hypernetworks.
\newblock \emph{International Conference on Learning Representations (ICLR)},
  2017.

\bibitem[Hong et~al.(2019)Hong, Liu, and Yuan]{hong2019dnnvolvis}
Fan Hong, Can Liu, and Xiaoru Yuan.
\newblock {DNN-VolVis}: Interactive volume visualization supported by deep
  neural network.
\newblock In \emph{IEEE Pacific Visualization Symposium (PacificVis)}, pp.\
  282--291, 2019.
\newblock \doi{10.1109/PacificVis.2019.00041}.

\bibitem[Jeong et~al.(2024)Jeong, Li, Johnson, Liu, and
  Berger]{jeong2024texttf}
Sangwon Jeong, Jixian Li, Christopher Johnson, Shusen Liu, and Matthew Berger.
\newblock Text-based transfer function design for semantic volume rendering,
  2024.
\newblock arXiv:2406.15634.

\bibitem[Jiang et~al.(2026)Jiang, Liu, Yang, Di, and Jin]{jiang2026particlegs}
Bo~Jiang, Youyuan Liu, Taolue Yang, Sheng Di, and Sian Jin.
\newblock {3D} gaussian splatting for scientific particle data compression and
  rendering, 2026.
\newblock arXiv:2607.22956.

\bibitem[Kerbl et~al.(2023)Kerbl, Kopanas, Leimk{\"u}hler, and Drettakis]{3dgs}
Bernhard Kerbl, Georgios Kopanas, Thomas Leimk{\"u}hler, and George Drettakis.
\newblock 3d gaussian splatting for real-time radiance field rendering.
\newblock \emph{ACM Transactions on Graphics}, 42\penalty0 (4):\penalty0
  139:1--139:14, 2023.
\newblock \doi{10.1145/3592433}.

\bibitem[Kniss et~al.(2002)Kniss, Kindlmann, and
  Hansen]{kniss2002multidimensional}
Joe Kniss, Gordon Kindlmann, and Charles Hansen.
\newblock Multidimensional transfer functions for interactive volume rendering.
\newblock \emph{IEEE Transactions on Visualization and Computer Graphics},
  8\penalty0 (3):\penalty0 270--285, 2002.
\newblock \doi{10.1109/TVCG.2002.1021579}.

\bibitem[Li et~al.(2026)Li, Perera, Forbes, Moreland, Pugmire, Klasky, Chao,
  and Shen]{li2026gssurrogate}
Ziwei Li, Rumali Perera, Angus Forbes, Ken Moreland, Dave Pugmire, Scott
  Klasky, Wei-Lun Chao, and Han-Wei Shen.
\newblock {GS-Surrogate}: Deformable gaussian splatting for parameter space
  exploration of ensemble simulations, 2026.
\newblock arXiv:2604.06358.

\bibitem[Liu et~al.(2026)Liu, Gao, Planche, Chen, Nguyen, Zheng, Choudhuri,
  Chen, Wang, Feng, and Wu]{liu2025universal}
Rong Liu, Zhongpai Gao, Benjamin Planche, Meida Chen, Van~Nguyen Nguyen, Meng
  Zheng, Anwesa Choudhuri, Terrence Chen, Yue Wang, Andrew Feng, and Ziyan Wu.
\newblock Universal beta splatting.
\newblock In \emph{ICLR}, 2026.
\newblock arXiv:2510.03312.

\bibitem[Lu et~al.(2021)Lu, Jin, Pang, Zhang, and Karniadakis]{lu2021deeponet}
Lu~Lu, Pengzhan Jin, Guofei Pang, Zhongqiang Zhang, and George~Em Karniadakis.
\newblock Learning nonlinear operators via {DeepONet} based on the universal
  approximation theorem of operators.
\newblock \emph{Nature Machine Intelligence}, 3:\penalty0 218--229, 2021.
\newblock \doi{10.1038/s42256-021-00302-5}.

\bibitem[Niedermayr et~al.(2024)Niedermayr, Neuhauser, Petkov, Engel, and
  Westermann]{niedermayr2024application}
Simon Niedermayr, Christoph Neuhauser, Kaloian Petkov, Klaus Engel, and
  R{\"u}diger Westermann.
\newblock Application of {3D} {G}aussian splatting for cinematic anatomy on
  consumer class devices.
\newblock In \emph{Vision, Modeling, and Visualization (VMV)}. The Eurographics
  Association, 2024.
\newblock arXiv:2404.11285.

\bibitem[Tang \& Wang(2026)Tang and Wang]{tang2026econgs}
Kaiyuan Tang and Chaoli Wang.
\newblock {ECoNGS}: Efficient compressive neural gaussian splats for volume
  visualization, 2026.
\newblock arXiv:2607.18466.

\bibitem[Tang et~al.(2025)Tang, Yao, and Wang]{tang2025ivrgs}
Kaiyuan Tang, Siyuan Yao, and Chaoli Wang.
\newblock {iVR-GS}: Inverse volume rendering for explorable visualization via
  editable {3D} gaussian splatting.
\newblock \emph{IEEE Transactions on Visualization and Computer Graphics},
  31\penalty0 (6):\penalty0 3783--3795, 2025.
\newblock \doi{10.1109/TVCG.2025.3567121}.

\bibitem[Wang et~al.(2004)Wang, Bovik, Sheikh, and Simoncelli]{wang2004ssim}
Zhou Wang, Alan~C. Bovik, Hamid~R. Sheikh, and Eero~P. Simoncelli.
\newblock Image quality assessment: From error visibility to structural
  similarity.
\newblock \emph{IEEE Transactions on Image Processing}, 13\penalty0
  (4):\penalty0 600--612, 2004.
\newblock \doi{10.1109/TIP.2003.819861}.

\bibitem[Weiss \& Westermann(2022)Weiss and Westermann]{weiss2022diffdvr}
Sebastian Weiss and R{\"u}diger Westermann.
\newblock Differentiable direct volume rendering.
\newblock \emph{IEEE Transactions on Visualization and Computer Graphics},
  28\penalty0 (1):\penalty0 562--572, 2022.
\newblock \doi{10.1109/TVCG.2021.3114769}.

\bibitem[Yao et~al.(2025)Yao, Lu, and Wang]{yao2025visnerf}
Siyuan Yao, Yunfei Lu, and Chaoli Wang.
\newblock {ViSNeRF}: Efficient multidimensional neural radiance field
  representation for visualization synthesis of dynamic volumetric scenes.
\newblock In \emph{IEEE Pacific Visualization Conference (PacificVis)}, pp.\
  235--245, 2025.
\newblock \doi{10.1109/PacificVis64226.2025.00029}.

\bibitem[Yu et~al.(2024)Yu, Chen, Huang, Sattler, and
  Geiger]{yu2023mipsplattingaliasfree3dgaussian}
Zehao Yu, Anpei Chen, Binbin Huang, Torsten Sattler, and Andreas Geiger.
\newblock Mip-splatting: Alias-free 3d gaussian splatting.
\newblock In \emph{CVPR}, 2024.

\bibitem[Zaheer et~al.(2017)Zaheer, Kottur, Ravanbakhsh, P{\'o}czos,
  Salakhutdinov, and Smola]{zaheer2017deepsets}
Manzil Zaheer, Satwik Kottur, Siamak Ravanbakhsh, Barnab{\'a}s P{\'o}czos,
  Ruslan Salakhutdinov, and Alexander~J. Smola.
\newblock Deep sets.
\newblock In \emph{Advances in Neural Information Processing Systems},
  volume~30, 2017.

\bibitem[Zwicker et~al.(2001)Zwicker, Pfister, Van~Baar, and
  Gross]{zwicker2001ewa}
Matthias Zwicker, Hanspeter Pfister, Jeroen Van~Baar, and Markus Gross.
\newblock Ewa volume splatting.
\newblock In \emph{Proceedings Visualization, 2001. VIS'01.}, pp.\  29--36.
  IEEE, 2001.
\newblock \doi{10.1109/VISUAL.2001.964490}.

\end{thebibliography}
